\documentclass[10pt,twocolumn,letterpaper]{article}

\usepackage{ijcb}

\usepackage{times}
\usepackage{booktabs}
\usepackage{epsfig}
\usepackage{graphicx}
\usepackage{amsmath}
\usepackage{amssymb}
\usepackage{tabularx}
\usepackage{multirow}
\usepackage{caption}
\usepackage{array}
\usepackage{subcaption}
\usepackage{changepage}
\usepackage{xcolor}
\usepackage{url}

\newcolumntype{P}[1]{>{\centering\arraybackslash}p{#1}}
\newcolumntype{M}[1]{>{\centering\arraybackslash}m{#1}}

\ijcbfinalcopy 

\def\ijcbPaperID{138} 

\ifijcbfinal\fi

\makeatletter
\def\ps@IEEEtitlepagestyle{
\def\@oddfoot{\mycopyrightnotice}
\def\@evenfoot{}
}
\def\mycopyrightnotice{
{\hfill \footnotesize 978-1-6654-3780-6/21/\$31.00 \copyright 2021 IEEE\hfill}
}
\makeatother

\begin{document}

\title{A Unified Model for Fingerprint Authentication \\and Presentation Attack Detection}

\author{Additya Popli\thanks{These authors have contributed equally.} \textsuperscript{1}, Saraansh Tandon\footnotemark[1] \textsuperscript{1}, Joshua J. Engelsma\textsuperscript{2}, Naoyuki Onoe\textsuperscript{3}, Atsushi Okubo\textsuperscript{4},\\ and Anoop Namboodiri\textsuperscript{1}\\
\textsuperscript{1}IIIT Hyderabad, India \quad
\textsuperscript{2}Michigan State University, USA \quad 
\textsuperscript{3}Sony Research India Pvt. Ltd., India \quad \\
\textsuperscript{4}Sony Group Corporation, Japan\\
{\tt\small  \{additya.popli, saraansh.tandon\}@research.iiit.ac.in},\\
{\tt\small engelsm7@msu.edu, \{naoyuki.onoe, atsushi.okubo\}@sony.com, anoop@iiit.ac.in}
}

\maketitle
\thispagestyle{empty}

\begin{abstract}
   Typical fingerprint recognition systems are comprised of a spoof detection module and a subsequent recognition module, running one after the other. In this paper, we reformulate the workings of a typical fingerprint recognition system. In particular, we posit that both spoof detection and fingerprint recognition are correlated tasks. Therefore, rather than performing the two tasks separately, we propose a joint model for spoof detection and matching\footnote{The terms authentication, verification and matching have been used interchangeably to refer to a 1:1 match surpassing a threshold.} to simultaneously perform both tasks without compromising the accuracy of either task. We demonstrate the capability of our joint model to obtain an authentication accuracy (1:1 matching) of TAR = 100\% @ FAR = 0.1\% on the FVC 2006 DB2A dataset while achieving a spoof detection ACE of 1.44\% on the LiveDet 2015 dataset, both maintaining the performance of stand-alone methods. In practice, this reduces the time and memory requirements of the fingerprint recognition system by 50\% and 40\%, respectively; a significant advantage for recognition systems running on resource-constrained devices and communication channels. The project page for our work is available at: \url{bit.ly/ijcb2021-unified}.
\end{abstract}
\vspace{-8mm}

\let\thefootnote\relax\footnotetext{\mycopyrightnotice}

\section{Introduction}

\begin{figure}[h]
  \centering
  \includegraphics[width=\columnwidth]{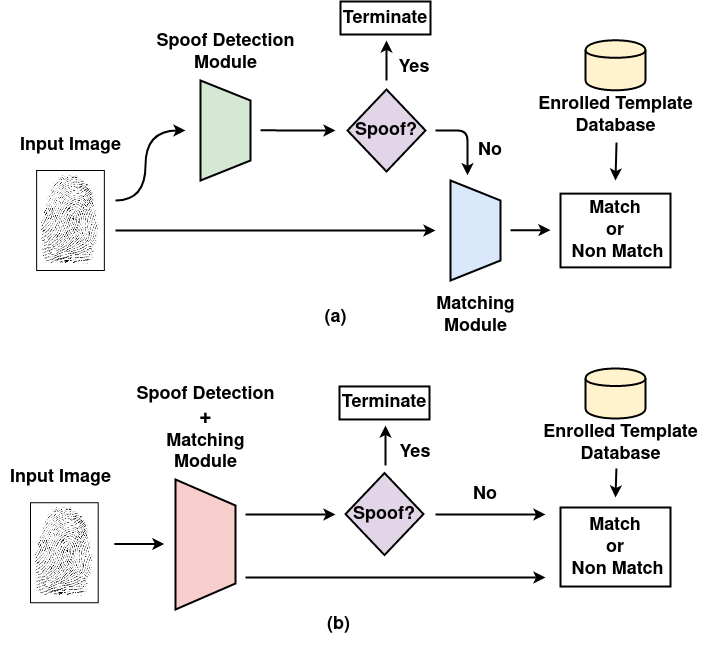}
  \caption{(a) Standard fingerprint recognition pipeline with separate networks running either in parallel (larger memory requirement) or series (increased time) for fingerprint spoof detection and matching (b) Proposed pipeline with a common network for both the tasks that reduces both time and memory consumption without significantly affecting the accuracy of either task.}
  \label{fig:introduction}
  \vspace{-4.5mm}
\end{figure}

\begin{figure}[h]
\centering
\begin{subfigure}{.4\columnwidth}
  \centering
  \includegraphics[width=\columnwidth]{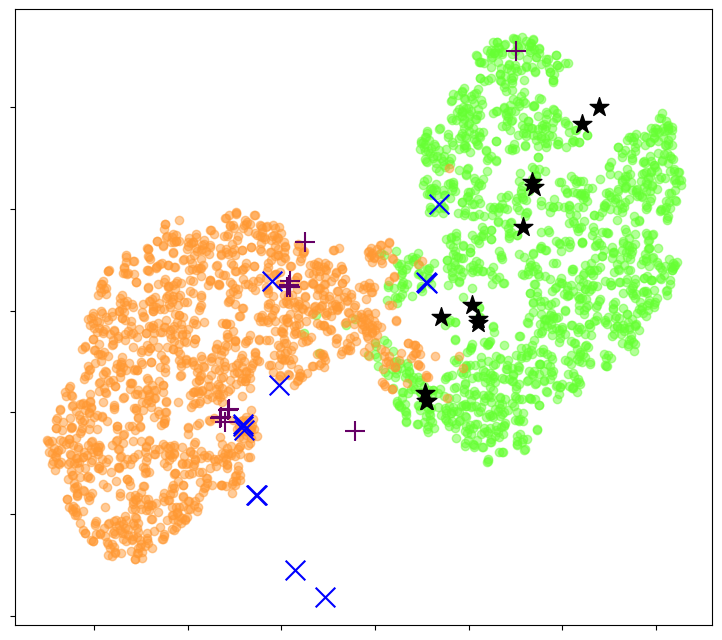}
  \caption{Model trained for spoof detection only}
  \label{plot1}
\end{subfigure}
\hspace{1em}
\begin{subfigure}{.4\columnwidth}
  \centering
  \includegraphics[width=\columnwidth]{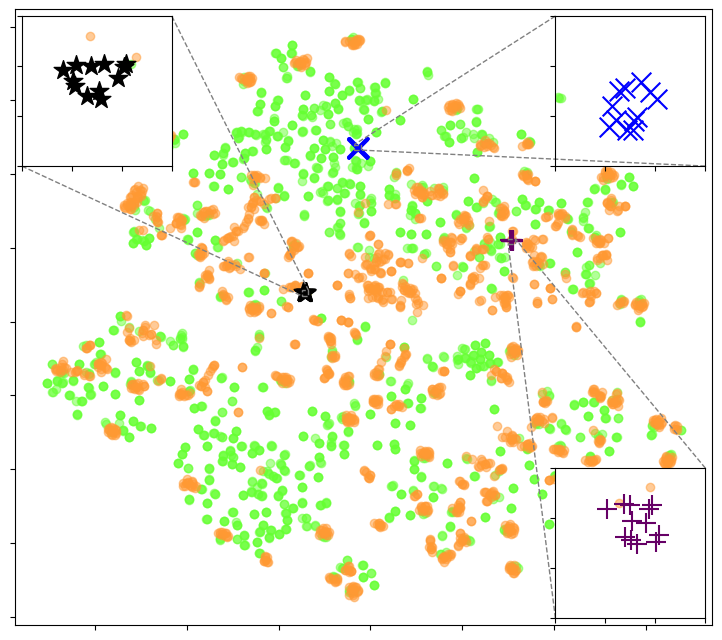}
  \caption{Model trained for matching only}
  \label{plot2}
\end{subfigure}

\begin{subfigure}{.4\columnwidth}
  \centering
  \includegraphics[width=\columnwidth]{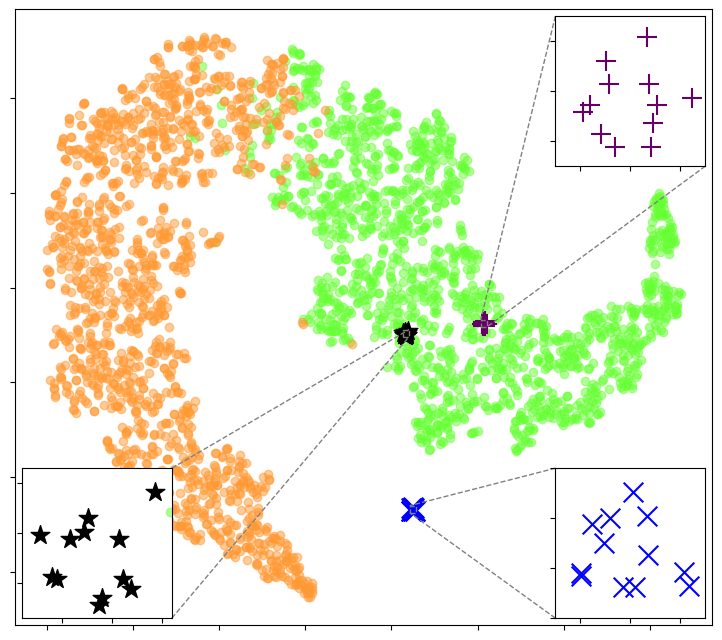}
  \caption{Proposed Approach}
  \label{plot3}
\end{subfigure}
\hspace{1em}
\begin{subfigure}{.4\columnwidth}
  \centering
  \includegraphics[width=0.75\columnwidth]{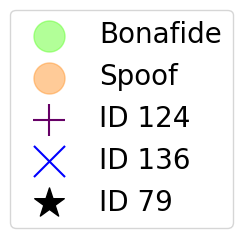}
  \caption{Legend}
  \label{plot4}
\end{subfigure}
\caption{t-SNE visualization of the embeddings generated by models trained for spoof detection and fingerprint matching. Unlike (a) and (b) which only extract embeddings useful for their respective task (\textit{i.e.} spoof detection, or matching), our model (c) is able to extract embeddings which can be used for both tasks.}
\vspace{-0.5cm}
\label{fig:embedding_plot}
\end{figure}

Due to their widespread usage in many different applications, fingerprint recognition systems are a prime target for attackers. One of the most widely known methods of attack is known as a presentation attack (PA), which can be realized through the use of commonly available materials like gelatin, play-doh and silicone or more expensive and sophisticated 3D printing techniques (these subset of presentation attacks are also known as spoof attacks). To counter these attacks, various fingerprint presentation attack detection (FPAD) approaches to automatically detect and flag spoof attacks prior to performing authentication have been proposed~\cite{marasco2014survey}.

The typical pipeline of a fingerprint recognition system equipped with such a spoof detector is illustrated in Figure~\ref{fig:introduction}a. First, a fingerprint image acquired by the fingerprint reader is passed to a spoof detector, where a decision is made as to whether the image comes from a live (bonafide) fingerprint or from a fake (spoof) fingerprint. If the image is determined to be bonafide, it is passed on to the authentication module for matching. In such a system, the spoof detector can run in parallel with the authentication module (only releasing the decision if the spoof detector decides bonafide), or the spoof detector can run in series with the authentication module (as a pre-filter before ever running authentication). Both modules running in parallel requires significant memory, especially for recent spoof detection and matching algorithms that leverage deep convolutional neural networks (CNNs). If instead the algorithms were to run in series, memory can be saved, but the time needed to release a decision to the user is increased. Both of these limitations are most prominently manifested on resource constrained devices and weak communication channels.
Given these limitations, we propose a reformulation to the typical workings of a fingerprint recognition system (Figure~\ref{fig:introduction}a). We posit that FPAD and fingerprint matching are correlated tasks and propose to train a model that is able to jointly perform both functions. Our system design is shown in Figure~\ref{fig:introduction}b. Our joint model maintains the accuracy of published standalone FPAD and matching modules, while requiring 50\% and 40\% less time and memory, respectively. 

Our motivation for coupling  FPAD and fingerprint matching into a single model and our intuition for doing so are based upon the following observation. Many FPAD algorithms and most fingerprint matching algorithms rely heavily on Level 1 (ridge-flow), Level 2 (minutiae) and Level 3 (pores) features. For example, minutiae points have been shown to provide significant utility in both fingerprint matching \cite{Cappelli2010MCC}, and FPAD~\cite{chugh2018fingerprint} systems. Fingerprint pores have also been used for both tasks~\cite{marcialis2010analysis, jain2007pores}.

Given the possibility of correlation between FPAD and fingerprint matching tasks (\textit{i.e.} they both benefit by designing algorithms around similar feature sets), we conduct a study in this paper to closely examine the relationship between these two tasks with the practical benefit of reducing the memory and time consumption of fingerprint recognition systems. Our work is also motivated by similar work in the face recognition and face PAD domains where the correlation of these two tasks was demonstrated via a single deep network for both tasks~\cite{ying2018liveface}. However, to the best of our knowledge, this is the first such work to investigate and show that PAD and recognition are related in the fingerprint domain where different feature sets are exploited than in the face domain.

More concisely, our contributions are: \vspace{-2mm}
\begin{itemize}
    \item A study to examine the relationship between fingerprint matching and FPAD.
    We show that features extracted from a state-of-the-art fingerprint matcher (\cite{engelsma2019learning}) can also be used for spoof detection. This serves as the motivation to build a joint model for both the tasks. \vspace{-2mm}
    \item The first model capable of simultaneously performing FPAD and fingerprint matching. Figure~\ref{fig:embedding_plot} shows that the embeddings extracted from this model for bonafide and spoof images are well separated while keeping the distance between embeddings extracted from different impressions of the same fingerprint together. \vspace{-2mm}
    \item Experimental results demonstrating matching accuracy of TAR = 100\% @ FAR = 0.1\% on FVC 2006 and fingerprint presentation attack detection ACE of 1.44\% on LiveDet 2015,  both similar to the performance of individual published methods. \vspace{-2mm}
    \item A reduction in time and memory requirements for a fingerprint recognition system of 50\% and 40\%, respectively, without sacrificing significant system accuracy. Our algorithm has significant advantages for resource constrained fingerprint recognition systems, such as those running on smartphone devices.
\end{itemize}

\section{Related Work}

\subsection{Fingerprint Spoof Detection}

Fingerprint spoof detection approaches that have been proposed in the literature can be broadly classified into hardware-based and software-based solutions~\cite{marasco2014survey}. While hardware-based solutions rely upon detecting the physical characteristics of a human finger with the help of additional sensor(s), software-based approaches extract features from fingerprint images already captured for the purpose of fingerprint matching and do not require any additional sensors~\cite{marasco2014survey}. Early software-based approaches relied on hand-crafted or engineered features extracted from the fingerprint images to classify them as either bonafide or spoofs~\cite{marasco2014survey}. However, more recent approaches are based upon deep Convolutional Neural Networks (CNN), and have been shown to significantly outperform previous approaches. The current state-of-the-art CNN based method~\cite{chugh2018fingerprint} utilized multiple local patches of varying resolutions centered and aligned around fingerprint minutiae to train two-class MobileNet-v1~\cite{howard2017mobilenets} and Inception v3~\cite{Szegedy2016RethinkingTI} networks. 

We posit that another important limitation of many state-of-the-art FPAD algorithms that needs to be further addressed is that of their computational complexity. In particular, the recent CNN based FPAD algorithms are memory and processor intensive algorithms (\textit{e.g.} the Inception v3 architecture utilized by several state-of-the-art approaches~\cite{chugh2018fingerprint} is comprised of 27M parameters and requires 104 MB of memory) that will cause challenges on resource constrained devices such as smartphones. Even ``lighter weight" CNN models such as MobileNet (used in~\cite{chugh2018fingerprint}) will also add computational complexity to a resource constrained device, particularly when utilized in line with a current fingerprint recognition pipeline (first perform spoof detection, then perform matching). Therefore, in this work, we aim to alleviate some of the computational burden of the FPAD module in a fingerprint recognition system via a joint model which performs both fingerprint matching and FPAD. 

\begin{figure*}[h]
  \centering
  \includegraphics[width=0.9\textwidth]{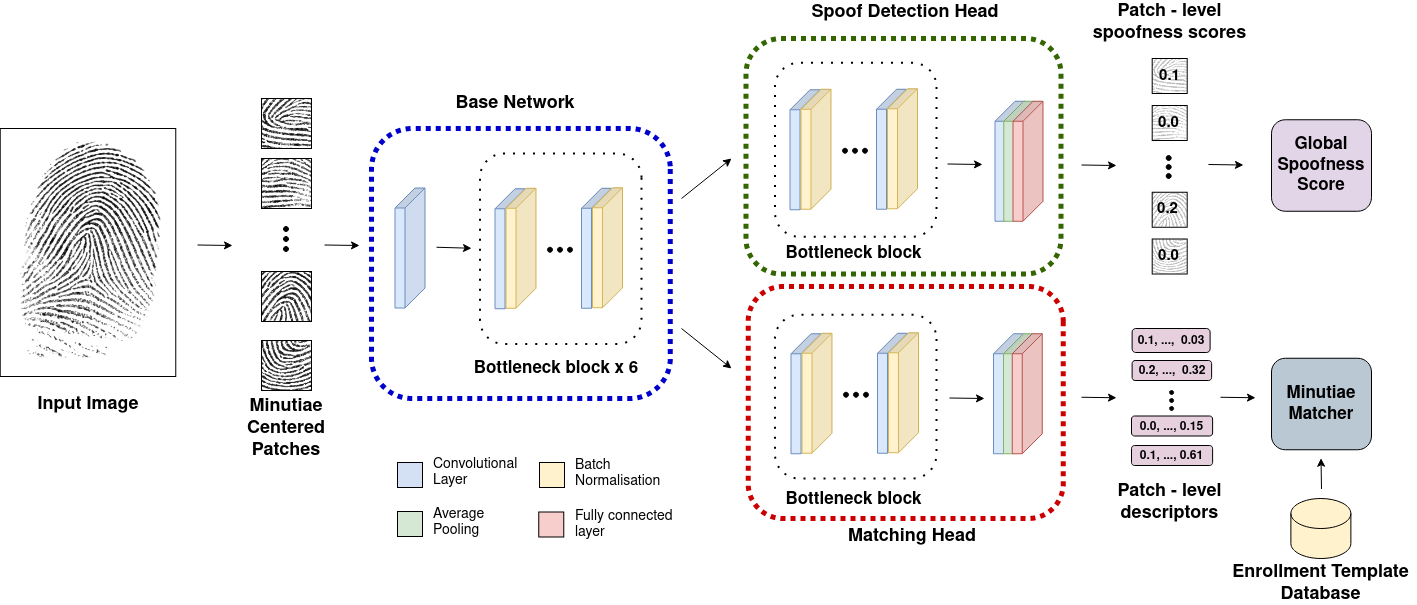}
  \caption{Proposed architecture with $split\_point=1$. Minutiae-centered patches of size 96 $\times$ 96 are extracted from the input image and fed to the base network which extracts a common feature map. This feature map is then fed to the spoof detection and matching heads to obtain patch-level spoofness scores and minutiae descriptors. The patch level scores are averaged to obtain an image level spoofness score and the patch-level descriptors from two different fingerprints are fed to the matching algorithm of \cite{jain2018endtoend} to obtain a similarity score. }
  \label{fig:architecture}
\end{figure*}

\subsection{Fingerprint Matching}

A plethora of work has been done in the area of fingerprint matching. In our work, we are primarily interested in deep-learning based solutions since they (i) have shown to outperform traditional solutions for fingerprint spoof detection \cite{chugh2018fingerprint} and on some databases, even for fingerprint matching \cite{jain2018endtoend, engelsma2019learning} and (ii) can be easily modified for simultaneous matching and spoof detection via a shared architecture and modified loss function.

One of the most well known hand-crafted approaches for fingerprint recognition is the Minutiae Cylinder Code (MCC)~\cite{Cappelli2010MCC}. More recently, CNNs have been used to extract fixed-length representations of fingerprints~\cite{engelsma2019learning} and also for learning local minutiae descriptors~\cite{jain2018endtoend}. In particular, the authors in~\cite{engelsma2019learning} proposed a network called DeepPrint which demonstrated high-levels of matching accuracy on several benchmark datasets. The limitation of DeepPrint's fixed-length representation is that it may fail when the probe and enrollment fingerprint are comprised of significantly non-overlapping portions of the fingerprint. In~\cite{jain2018endtoend} a CNN was used to extract local minutiae descriptors to boost latent fingerprint matching accuracy. Given the high accuracy and open source code provided by~\cite{jain2018endtoend}, we incorporate their approach into our joint model.

\begin{table}[h]
\centering
\footnotesize
\caption{FPAD on LivDet 15 \cite{mura2015livdet} using feature maps from a DeepPrint~\cite{engelsma2019learning} student model (matcher) as input.}
{\renewcommand{\arraystretch}{1}%
\begin{tabular}{  P{1.25cm} P{1cm} P{1cm} P{1cm} P{1cm} P{0.75cm} }
        \toprule
        \textbf{Sensor} & Cross Match & Digital Persona & Green Bit & HiScan & \textbf{Avg.}\\[0.2cm]
        \midrule
        \textbf{ACE (\%)} & 7.43 & 13.48 & 6.87 & 10.2 & \textbf{9.49}\\[0.2cm]
        \bottomrule
\end{tabular}}
\flushleft{\footnotesize{}}
\label{table: motivation}
\vspace{-5mm}
\end{table}

\section{Motivation}
\label{sec:motivation}

As an initial experiment to check whether there exists a relationship between fingerprint matching and spoof detection, we build a spoof detection network on top of the features extracted by DeepPrint~\cite{engelsma2019learning}. Specifically, we use the Inception-v3 \cite{Szegedy2016RethinkingTI} network as student model\footnote{We have used a student model since the data used in the original paper is not publicly available and the authors of \cite{engelsma2019learning} agreed to share the weights of their model to use it as a teacher network.} of DeepPrint by training it to extract embeddings from fingerprints as close as possible to the embeddings extracted by DeepPrint for the same input images obtained from the NIST SD-300 and NIST SD-302 datasets. Next, we use the intermediate feature maps extracted from the \textit{Mixed\_5d} layer of our student model as inputs for a shallow spoof detection network (refer 
Appendix \ref{sec:motivation_architecture} 
for architecture details). We observe that this spoof detection model is able to classify these intermediate identity related feature maps as live or spoof with high accuracy (Table \ref{table: motivation}). It is important to note that our shallow spoof detection model is trained on features maps of size $35 \times 35$ which are much smaller than the original student model input size (of the fingerprint image) of $448 \times 448$.

After establishing (experimentally) that common features can be used for both fingerprint matching and spoof detection, we focus our efforts on a single, joint-model that is capable of performing both these tasks while maintaining the accuracy of published stand-alone approaches.

\section{Methodology}

To train a single model capable of both fingerprint spoof detection and fingerprint matching, we build a multi-branch CNN called DualHeadMobileNet (Figure~\ref{fig:architecture}). The input to our model is a $96\times~96$ minutiae centered patch (resized to $224\times~224$ according to the model input size). One branch of the CNN performs the task of spoof detection (outputting ``spoofness" scores between 0 and 1 for the patch), while the other branch extracts a local minutiae descriptor of dimension 64 from the patch (for the purpose of matching). Both branches share a common stem network. This parameter sharing is what enables us to reduce the time and memory constraints of our fingerprint recognition system. Rather than training two networks for two tasks, we share a number of parameters in the stem (given our hypothesis that these two tasks are related and can share a number of common parameters). To obtain a final spoof detection decision, we average the ``spoofness" scores of all the minutiae-centered patches extracted from a given fingerprint. To conduct matching, we aggregate all of the minutiae descriptors extracted from all the minutiae-centered patches in a given fingerprint image and subsequently feed them to the minutiae matcher open sourced in~\cite{jain2018endtoend}.

\subsection{DualHeadMobileNet (DHM)} \label{section:DualHeadMobileNet}

Given the success of~\cite{chugh2018fingerprint} in using MobileNet-v1 for spoof detection, we have used MobileNet-v2 \cite{Sandler2018MobileNetV2IR} as the starting point for our DualHeadMobileNet. We have chosen MobileNet-v2 over MobileNet-v1 \cite{howard2017mobilenets} because it (i) obtains higher classification accuracy on benchmark datasets (ii) uses 2 times fewer operations and (iii) needs 30 percent fewer parameters as compared to MobileNet-v1. Since we are highly cognizant of computational complexity, less parameters is a significant motivating factor in our selection.

We modify the MobileNet-v2 architecture for our experiments in the following manner:

\begin{itemize}
    \item{\textbf{Base Network:}} This is a sub-network which is used to generate a common feature map for both spoof detection and minutiae descriptor extraction. It consists of the initial layers of MobileNet-v2 depending on the $split\_point$ (further explained later).
    
    \item{\textbf{Spoof Detection Head:}} This is a specialized sub-network trained for spoof detection and consists of the remaining layers (layers not included in the base network) of MobileNet-v2. The final linear layer is modified to obtain only two outputs (\textit{i.e.} live or spoof).

    \item{\textbf{Matching Head:}} This is a specialized sub-network trained for fingerprint matching and is identical to the spoof detection head except that the final linear layer here is modified to obtain a feature vector of the required embedding size. This feature vector is a minutiae descriptor of the input minutiae patch.
\end{itemize} 

To indicate the number of bottleneck blocks in the base-network and the separate head networks, we use the variable $split\_point$ where a network with $split\_point$ = $x$ means that $x$ bottleneck blocks are used in each head network, and 7 - $x$ bottleneck blocks are present in the base network. The complete model architecture is shown in Figure~\ref{fig:architecture}.

Please refer to
Appendix \ref{sec:robustness_architecture} 
for experiments using other networks as backbone architectures.

\subsection{Joint Training} \label{joint_training}

The two heads are trained separately to optimize the loss for their specific tasks while the common base network is trained to optimize a weighted sum of these individual losses, as shown in Equation~\ref{equation: joint_loss} where $w_{sd}$ and $w_{m}$ are the weights for the spoof detection and matching loss.

\begin{equation} \label{equation: joint_loss}
    \mathcal{L}_{total} = w_m \mathcal{L}_m + w_{sd} \mathcal{L}_{sd}
\end{equation}

The spoof detection head is trained via a cross-entropy loss ($\mathcal{L}_{sd}$) based on the ground truth of the input minutiae patch (\textit{i.e.} if the input patch is a live or spoof). Meanwhile the matching head outputs descriptors which are regressed to ground truth minutiae descriptors using an L2-norm ($\mathcal{L}_{m}$). These ground truth minutiae descriptors are extracted from the minutiae patch using the algorithm in~\cite{jain2018endtoend}, \textit{i.e.} the model in~\cite{jain2018endtoend} is a teacher model which our DHM seeks to mimic as a student. Due to the non-availability of a large and patch-wise labelled public dataset we use a student-teacher framework which eliminates the need of identity labels during the training procedure.

\section{Experiments and Results}

After training our DHM, we conduct experiments to demonstrate the capability of our joint model to accurately perform both spoof detection and matching. In doing so, we demonstrate that spoof detection and fingerprint matching are indeed correlated tasks. Please refer to 
Appendix \ref{sec:hyperparams}
for implementation and hyperparameter details.

\subsection{Datasets}
For spoof detection, we report on the LivDet 2015 \cite{mura2015livdet} and LivDet 2017 \cite{livdet17} datasets. For fingerprint matching, we report on the FVC 2000 \cite{fvc2000}, FVC 2002 \cite{fvc2002}, FVC 2004 \cite{fvc2004} and FVC 2006 \cite{fvc2006} datasets following the official protocols. Each sensor of the FVC 2002 and 2004 datasets contains 800 fingerprint images (100 fingers $\times$ 8 impressions), leading to 2800 genuine 
($ = \frac{8 \times 7}{2} \times 100$) 
and 4950 imposter 
($ = \frac{100 \times 99}{2}$) 
comparisons, while each sensor within the FVC 2006 dataset contains 1680 images (140 fingers $\times$ 12 impressions) leading to 9240 
($ = \frac{12 \times 11}{2} \times 140$) 
genuine and 9730 
($ = \frac{140 \times 139}{2}$) 
imposter comparisons.

The Orcanthus sensor of the LivDet 2017 dataset and non-optical sensors of the FVC datasets have been excluded for evaluation since we want to ensure that the sensor used to acquire our training images (\textit{e.g.} an optical sensor from LiveDet training partition) uses similar sensing technology as the sensor used to acquire our testing images (\textit{e.g.} an optical sensor from FVC). Therefore, we utilize only those subsets of data from LiveDet and FVC which both uses optical sensing. A future improvement to this work could be to make our model invariant to the differences in sensor characteristics observed across different sensing technologies.  

\subsection{Comparison with state-of-the-art methods}

\subsubsection{Spoof Detection} \label{mfsb}
We use our implementation of Fingerprint Spoof Buster~\cite{chugh2018fingerprint} (current SOTA) as a FPAD baseline. Since the authors did not provide official code or models, we train our own models in order to present a fair comparison both in terms of accuracy and system constraints (refer to Section~\ref{section:constraints}). For the reasons mentioned in Section~\ref{section:DualHeadMobileNet}, we use MobileNet-v2 (unless specified otherwise) instead of the MobileNet-v1 network. Also, since our goal is to develop a fast and memory-efficient system, we only train a single model using patches of size 96 $\times$ 96, as opposed to an ensemble of models trained on different patch sizes (as done by the authors in \cite{chugh2018fingerprint}). This baseline (referred to as $m$-FSB) achieves an average classification error of 1.48\% on the LivDet 2015 dataset \cite{mura2015livdet} compared to 0.98\% reported in \cite{chugh2018fingerprint} (using an ensemble of models) and even outperforms the original model on two of the four sensors present in the dataset.

Table \ref{table: main_table_sd}, shows the spoof detection results of our re-implementation of the baseline~\cite{chugh2018fingerprint} as well as our joint model. The proposed joint model achieves comparable results to the baseline stand-alone spoof detector and even outperforms it on 4 out of the 6 sensors of the LivDet 2015 \cite{mura2015livdet} and LivDet 2017 \cite{livdet17} datasets. The joint model also outperforms the best performing algorithm of the LivDet 2017 Fingerprint Liveness Detection Competition (as per the results reported in \cite{livdet17}) on the Digital Persona sensor (ACE 3.4\% vs 4.29\%) and achieves similar performance on the GreenBit sensor (ACE 2.92\% vs 2.86\%).

\begin{table}[t]
\centering
\footnotesize
\caption{Spoof Detection Classification Error (ACE \%) \textsuperscript{$\dagger$}}
{\renewcommand{\arraystretch}{1}%
\begin{tabular}{  P{0.9cm} P{0.7cm} P{1cm} P{0.7cm} P{0.7cm} P{1cm} P{0.7cm} }
        \toprule
        
        \multirow{2}[5]{0.9cm}{\centering\textbf{Method}} & \multicolumn{4}{P{3.35cm}}{\textbf{LivDet 15}} & \multicolumn{2}{P{1.7cm}}{\textbf{LivDet 17}} \\[0.2cm]
        
        \cmidrule(r){2-5}\cmidrule(r){6-7} 
        
         & \textbf{Cross Match} & \textbf{Digital Persona} & \textbf{Green Bit} & \textbf{Hi Scan} & \textbf{Digital Persona} & \textbf{Green Bit} \\[0.2cm] 
        
        \midrule\midrule
        
        $m$-FSB &
        0.38 &
        \textbf{3.01} &
        0.5 &
        2.04 &
        3.51 &
        \textbf{2.63} \\[0.2cm]
        
        \midrule
        
        DHM &
        \textbf{0.28} &
        3.28 &
        \textbf{0.44} &
        \textbf{1.77} &
        \textbf{3.4} &
        2.92 \\[0.2cm]
        
        \bottomrule
\end{tabular}}
\flushleft{\footnotesize{
$m$-FSB refers to the baseline mentioned in Sec.~\ref{mfsb} and DHM refers to the DualHeadMobileNet network trained using Joint Training (Sec. \ref{joint_training}) with $split\_point$ = 0, $w_{sd}$ = 1 and $w_m$ = 10.
\\ \textsuperscript{$\dagger$} Please refer to Appendix \ref{sec:spoof_detection_long} for APCER and BPCER values.
}}
\label{table: main_table_sd}
\end{table}

\begin{table*}[t]
\centering
\footnotesize
\caption{Comparison of the matching performance (FRR \% @ FAR = 0.1\%) on FVC datasets}
{\renewcommand{\arraystretch}{0.75}%
\begin{tabular}{  P{2cm} P{2cm} P{2cm} P{1cm} P{1cm} P{1cm} P{1cm} P{1cm} P{1cm} P{1cm} }
        \toprule
        
        \multirow{2}[5]{2cm}{\centering\textbf{Method}} & \multirow{2}[5]{2cm}{\centering\textbf{Train Dataset}} &
        \multirow{2}[5]{2cm}{\centering\textbf{Train Sensor}} &
        \textbf{2006} & \multicolumn{2}{P{2cm}}{\textbf{2004}} & \multicolumn{2}{P{2cm}}{\textbf{2002}} & \multicolumn{2}{P{2cm}}{\textbf{2000}} \\[0.2cm]
        
        \cmidrule(r){4-4}\cmidrule(r){5-6}\cmidrule(r){7-8}\cmidrule(r){9-10}
        
        & & & \textbf{DB2A} & \textbf{DB1A} & \textbf{DB2A} & \textbf{DB1A} & \textbf{DB2A} & \textbf{DB1A} & \textbf{DB3A} \\[0.2cm]
        
        \midrule\midrule
        
        VeriFinger v11 & - & - & 0.00 & 2.86 & 3.01 & 0.11 & 0.07 & 0.11 & 1.04 \\[0.2cm]
        
        \midrule
        \midrule
        
        DeepPrint \cite{engelsma2019learning} & - & - & 0.32 & \textbf{2.43} & 5.93 & 7.61 & 10.32 & 4.57 & 6.50 \\[0.2cm]
        
        \midrule
        
        MCC \cite{Cappelli2010MCC} & - & - & 0.03 & 7.64 & \underline{5.6} & 1.57 & 0.71 & 1.86 & 2.43 \\[0.2cm]
        
        \midrule
        
        LatentAfis \cite{jain2018endtoend}\textsuperscript{$\ddagger$} & - & - & \textbf{0.00} & 5.64 & 7.17 & \underline{0.82} & \textbf{0.46} & 1.25 & \textbf{1.79} \\[0.2cm]
        
        \midrule\midrule
        
        \multirow{6}[15]{2cm}{\centering DHM\textsuperscript{$\dagger$}} & \multirow{4}[10]{2cm}{LivDet 2015} & CrossMatch & \textbf{0.00} & 5.89 & 6.99 & \textbf{0.75} & \underline{0.50} & \underline{1.04} & \textbf{1.79} \\[0.2cm]
        
        \cmidrule(r){3-10}
        
         &  & DigitalPersona & \textbf{0.00} & 6.36 & 6.81 & 1.14 & 0.61 & 1.14 & 2.36 \\[0.2cm]
        
        \cmidrule(r){3-10}
        
         &  & GreenBit & 0.02 & 7.25 & \textbf{5.34} & \underline{0.82} & \underline{0.50} & 1.11 & 2.14 \\[0.2cm]
        
        \cmidrule(r){3-10}
        
         &  & HiScan & \textbf{0.00} & 5.57 & 6.46 & \underline{0.82} & \underline{0.50} & \underline{1.04} & 2.21 \\[0.2cm]
        
        \cmidrule(r){2-10}\morecmidrules\cmidrule(r){2-10}
        
         & \multirow{2}[5]{2cm}{LivDet 2017} & DigitalPersona & \underline{0.01} & 6.32 & 7.03 & 0.93 & 0.54 & 1.32 & \underline{1.86} \\[0.2cm]
        
        \cmidrule(r){3-10}
        
         &  & GreenBit & \underline{0.01} & \underline{5.25} & 7.57 & 0.89 & \textbf{0.46} & \textbf{0.96} & 2.18 \\[0.2cm]
        
        \bottomrule
\end{tabular}}
\flushleft{\footnotesize{
\textsuperscript{$\dagger$} This refers to the DualHeadMobileNet network trained using joint training as described in Section \ref{joint_training} with $split\_point$ = 0, $w_{sd} = 1$ and $w_m$ = 10. \\
Best and second best results for the published matchers are in bold and underline respectively.
}}
\label{table: main_table_matching}
\end{table*}
 
\subsubsection{Matching}

We compare our results with the following baselines:
\begin{itemize}
    \item \textbf{MCC \cite{Cappelli2010MCC}}: We use VeriFinger SDK v11 (commercial SDK) for minutiae detection and the official SDK provided by the authors for matching.
    \item \textbf{DeepPrint \cite{engelsma2019learning}}: We use the weights provided by the authors for feature extraction and follow the official protocol for calculating the matching scores.
    \item \textbf{LatentAFIS \cite{jain2018endtoend}}: Since we use this method as a teacher for training our matching branch, we also compare our results with this method. We use the weights and the matching algorithm open-sourced\footnote{\url{https://github.com/prip-lab/MSU-LatentAFIS}} by the authors to obtain the matching score.
\end{itemize}
We also provide the matching results of a commercial fingerprint matcher (VeriFinger v11 SDK) for reference. While our joint model does not outperform VeriFinger on its own, we note that (i) our joint model does outperform all existing baselines taken from the open literature in many testing scenarios and (ii) the existing published baselines are also inferior to Verifinger in most testing scenarios. 

To best the performance of Verifinger, two problems would need to be solved: 1) We would need more discriminative minutiae descriptors than that which LatentAFIS currently provides as our ground-truth for DHM. Indeed LatentAFIS is somewhat of an upperbound for DHM in terms of matching since LatentAFIS is the teacher of DHM. 2) We would have to further improve the minutiae-matching algorithm of~\cite{jain2018endtoend}. While both of these tasks constitute interesting research, they fall outside of the scope of this paper which is focused on demonstrating the correlation of FPAD and matching and showing that the performance of existing systems for each task can be \textbf{\textit{maintained}} when combining both tasks into a single network - not on obtaining SOTA minutiae matching performance. By maintaining the matching accuracy of LatentAFIS~\cite{jain2018endtoend} with DHM with simultaneous FPAD, we have met these primary objectives.

It is important to note that the same joint models have been used for spoof detection (Table \ref{table: main_table_sd}) and matching (Table \ref{table: main_table_matching}), while the baselines are stand-alone models for their respective task. For example, the results corresponding to the fifth row in Table \ref{table: main_table_matching} and the first column (of the DHM) in Table \ref{table: main_table_sd} refer to the same model (weights) trained on the images of the CrossMatch sensor of the LivDet 15 \cite{mura2015livdet} dataset.

Please refer to Appendix \ref{sec:matching_feature_eval} for more experimental results.

\vspace{-1mm}
\subsection{Time and Memory} \label{section:constraints}
A fingerprint recognition system comprised of a ``front-end" spoof detector such as that of $m$-FSB and a fingerprint matcher such as that in~\cite{jain2018endtoend} requires 2.93 seconds of inference time per image\footnote{We ignore the time required for the minutiae matching since it is common to both the traditional and proposed pipeline and takes only $\sim$1ms.}, 1.46 seconds for spoof detection and 1.47 seconds for descriptor extraction for matching (assuming avg. number of minutiae of 49 as per the LivDet 15 dataset and batched processing) on an Intel Xeon E5-2640 v4 processor. In comparison, the \textbf{proposed approach consists of only a single network for both of the tasks and takes only 1.49 seconds, providing a speed up of 49.15\%}. The number of parameters and the memory requirements of the proposed approach are 2.7M and 10.38 MB compared to 4.5M and 17.29 MB of the traditional approach. 

Since this reduction in space and time consumption is most useful in resource constrained environments, we also benchmark the inference times of the traditional method and our proposed joint method on a OnePlus Nord mobile phone. While the traditional pipeline takes 62.18ms (per patch since mobile libraries do not support batch processing), the proposed approach takes only 32.03ms for feature extraction, resulting in a 48.49\% reduction in the feature extraction time. In this case, we first quantize the models (both baseline and joint) from 32-bits to 8-bits since quantization has shown to provide approximately 80\% speed-up without significant drop in accuracy \cite{chugh2018generalizationAndEfficiency}.

\section{Ablation Study}

All the ablation studies were done using the LivDet 2015 \cite{mura2015livdet} CrossMatch dataset (training split) for training. In Sections \ref{section: split} and \ref{section:suppress} we report the Spoof Detection ACE on LivDet 2015 \cite{mura2015livdet} CrossMatch testing dataset and Matching FRR (\%) @ FAR = 0.0\% on the FVC2006 \cite{fvc2006} DB2A dataset. Similar trends were observed with other datasets.

\begin{table}[h]
\centering
\footnotesize
\caption{Effect of varying the Split Point: similar performance is observed across different split points}
{\renewcommand{\arraystretch}{1}%
\begin{tabular}{ P{0.75cm} P{2.5cm} P{0.75cm} P{2cm}}
        \toprule
        
        \textbf{Split Point} & \textbf{Parameters (M) Base Network / Total} & \textbf{ACE (\%)} & \textbf{FRR (\%) @ \quad FAR = 0.0\%} \\[0.2cm]
        
        \midrule\midrule

        0 & 1.81 / 2.72 & 0.28 & 0.06 \\[0.1cm]
        
        \midrule
        
        1 & 1.34 / 3.19 & 0.41 & 0.18 \\[0.1cm]
        
        \midrule
        
        2 & 0.54 / 3.99 & 0.38 & 0.17 \\[0.1cm]
        
        \midrule
        
        3 & 0.24 / 4.29 & 0.41 & 0.04 \\[0.1cm]
        
        \bottomrule
        
\end{tabular}}
\flushleft{\footnotesize{}}
\label{table: split_point}
\vspace{-5mm}
\end{table}

\subsection{Effect of varying the split point} \label{section: split}
We vary the $split\_point$ of the DHM to further examine the correlation between fingerprint matching and spoof detection. A model with a higher $split\_point$ value indicates that the base network is shallower and hence the two specialized heads are more independent of each other. As shown in Table~\ref{table: split_point}, we notice that there is very little improvement for both spoof detection and matching even when we allow deeper specialised heads (\textit{i.e} increase the $split\_point$, and consequently the number of model parameters). This indicates that even when some bottleneck layers are allowed to train independently for the two tasks, they extract similar features. In a conventional fingerprint recognition pipeline, these redundant features would waste time and space, however our joint model eliminates these wasteful parameters by sharing them across both tasks.

\begin{table}[h]
\centering
\footnotesize
\caption{Effect of Suppression: suppressing one task negatively affects both the tasks }
{\renewcommand{\arraystretch}{1}%
\begin{tabular}{ P{2.25cm} P{1.25cm} P{3cm}}
        \toprule
        
        \textbf{Loss Suppressed} & \textbf{ACE (\%)} & \textbf{FRR (\%) @ FAR = 0.0\%} \\[0.2cm]
        
        \midrule\midrule

        None &  0.28 &	0.11 \\[0.1cm]
        
        \midrule
        
        Spoof Detection &  50	& 0.12\\[0.1cm]
        
        \midrule
        
        Matching & 0.47	& 98.06\\[0.1cm]
        
        \bottomrule
        
\end{tabular}}
\label{table: suppressed}
\end{table}

\subsection{Effect of Suppression} \label{section:suppress}
In order to better understand the degree of dependence between these two tasks, we try to train models to perform only one task (e.g matching) while suppressing any information which helps the other task (spoof detection, in this case). A decrease in matching performance due to suppression of spoof detection or vice-versa would indicate a strong dependence between the two tasks. In this case, the two model heads are trained similarly as in Joint Training (Section \ref{joint_training}), the only difference being in the merging of the two losses at the split point. In this case, the gradient flowing into the base network from the suppressed branch is first negated, and then added to the gradient flowing from the other branch, similar to the technique proposed in \cite{ganin2016dann} (diagram for the same is included in
Appendix \ref{sec:backprop_diagram}).

Although on suppressing spoof detection the matching error rate doesn't change much from 0.11\% to 0.12\% (refer Table \ref{table: suppressed}), suppressing matching increases the spoof detection ACE significantly from 0.28\% to 0.47\%. These findings again lend evidence to our hypothesis that spoof detection and fingerprint authentication are related tasks and as such can be readily combined into a joint model to eliminate wasteful memory and computational time.

\begin{figure}[h]
\begin{subfigure}{0.9\columnwidth}
  \centering
  \includegraphics[width=\columnwidth]{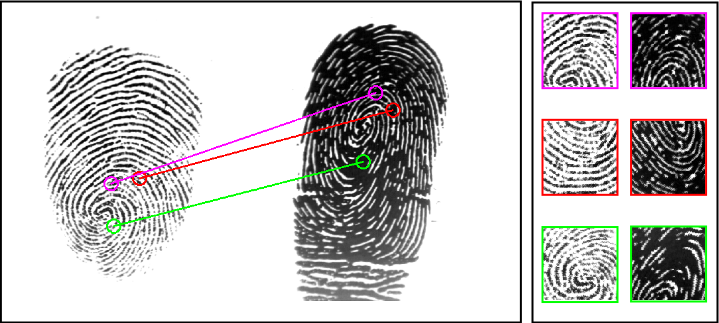}
  \caption{Genuine pair from FVC 2004 DB1A falsely rejected by the DHM. Matching minutiae pairs missed by the DHM have been marked.}
  \label{fig:false_rejects}
\end{subfigure}
\hspace{1em}
\begin{subfigure}{0.9\columnwidth}
  \centering
  \includegraphics[width=\columnwidth]{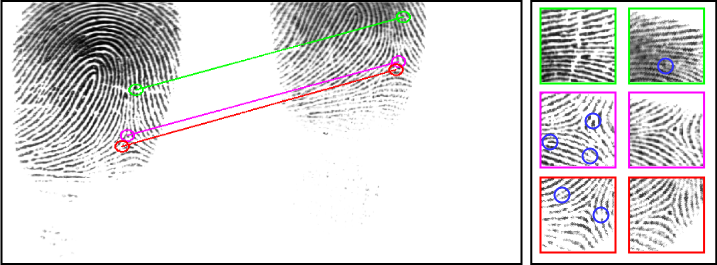}
  \caption{Imposter pair from FVC 2004 DB1A falsely accepted by the DHM. False minutiae pair correspondences have been marked on the left and minutiae points differentiating the falsely corresponding minutiae patches have been marked in blue (right).}
  \label{fig:false_accepts}
\end{subfigure}
\caption{Failure Analysis: In both Figures \ref{fig:false_rejects} and \ref{fig:false_accepts} three minutiae pairs and their corresponding patches have been highlighted with different colors for ease of viewing.}
\end{figure}

\section{Failure Cases}

We perform a qualitative analysis to understand the cases where our system fails. In particular, we focus on the fingerprint matching performance on the FVC 2004 DB1A dataset because our system (and also the baseline matching algorithm \cite{jain2018endtoend} on top of which be build our system) performs considerably worse than the best fingerprint matcher (DeepPrint \cite{engelsma2019learning}) for this particular database.

\subsection{False Rejects}
Figure~\ref{fig:false_rejects} shows a subset of mated-minutiae pairs (obtained using VeriFinger and manually verified) across two different acquisitions of the same finger which were rejected by DHM. Due to large distortion and difference in moisture, patches extracted around the same minutiae point (different acquisitions) are visually dissimilar and hence produce descriptors with low similarity. Global fingerprint matching techniques \cite{engelsma2019learning} which rely on the global ridge-flow information instead of local minutiae-patches are more robust to local deformations and hence perform better in such cases. 

\subsection{False Accepts} \label{section:false_accepts}
Figure~\ref{fig:false_accepts} shows acquisitions from two different fingers that are accepted as a genuine pair by our matcher. Although both the global fingerprint images and local minutiae patches look very similar and have some similar non-matching minutiae patches, there also are imposter minutiae patches (marked in non-blue) with spurious minutiae (marked in blue circles) that should be possible a network to differentiate. We believe that the network learns the general ridge-flow structure instead of specific differences like missing minutiae points and is hence unable to differentiate between the two patches. These cases we believe can be addressed by incorporating attention maps or minutiae-heat maps (as done by \cite{engelsma2019learning}) to guide the network to focus on more discriminative features in the minutiae descriptor. 

\section{Conclusion}

Existing fingerprint recognition pipelines consist of FPAD followed by a matching module. These two tasks are often treated independently using separate algorithms. Our experimental results indicate that these tasks are indeed related. In practice, this enables us to train a single joint model capable of performing FPAD and authentication at levels comparable to published stand-alone models while reducing the memory and time of the fingerprint recognition systems by 50\% and 40\%. We have also shown that our algorithm is applicable to patch based fingerprint recognition systems as well as full image recognition systems. In our ongoing research, we are investigating ways to further reduce the memory and computational complexity of fingerprint recognition systems, without sacrificing system accuracy. This will have tremendous benefit for fingerprint recognition systems running on resource constrained devices and communication channels.

{\small
\bibliographystyle{ieee}
\bibliography{egbib}

\begin{thebibliography}{10}\itemsep=-1pt

\bibitem{jain2018endtoend}
K.~Cao, D.-L. Nguyen, C.~Tymoszek, and A.~K. Jain.
\newblock End-to-end latent fingerprint search.
\newblock {\em Trans. Info. For. Sec.}, 15:880–894, Jan. 2020.

\bibitem{Cappelli2010MCC}
R.~{Cappelli}, M.~{Ferrara}, and D.~{Maltoni}.
\newblock Minutia cylinder-code: A new representation and matching technique
  for fingerprint recognition.
\newblock {\em IEEE Transactions on Pattern Analysis and Machine Intelligence},
  32(12):2128--2141, 2010.

\bibitem{chugh2018fingerprint}
T.~{Chugh}, K.~{Cao}, and A.~K. {Jain}.
\newblock Fingerprint spoof buster: Use of minutiae-centered patches.
\newblock {\em IEEE Trans. Info. For. Sec.}, 13(9):2190--2202, 2018.

\bibitem{chugh2018generalizationAndEfficiency}
T.~Chugh and A.~K. Jain.
\newblock Fingerprint presentation attack detection: Generalization and
  efficiency, 2018.

\bibitem{engelsma2019learning}
J.~J. {Engelsma}, K.~{Cao}, and A.~K. {Jain}.
\newblock Learning a fixed-length fingerprint representation.
\newblock {\em IEEE on Pattern Analysis and Machine Intelligence}, pages 1--1,
  2019.

\bibitem{Gajawada2019UniversalMT}
R.~{Gajawada}, A.~{Popli}, T.~{Chugh}, A.~{Namboodiri}, and A.~K. {Jain}.
\newblock Universal material translator: Towards spoof fingerprint
  generalization.
\newblock In {\em 2019 Int. Conf. on Biometrics (ICB)}, pages 1--8, 2019.

\bibitem{ganin2016dann}
Y.~Ganin and V.~Lempitsky.
\newblock Unsupervised domain adaptation by backpropagation.
\newblock In {\em Proceedings of the 32nd International Conference on Machine
  Learning - Volume 37}, ICML’15, page 1180–1189. JMLR.org, 2015.

\bibitem{resnet}
K.~{He}, X.~{Zhang}, S.~{Ren}, and J.~{Sun}.
\newblock Deep residual learning for image recognition.
\newblock In {\em 2016 IEEE Conference on Computer Vision and Pattern
  Recognition (CVPR)}, pages 770--778, 2016.

\bibitem{howard2017mobilenets}
A.~G. Howard, M.~Zhu, B.~Chen, D.~Kalenichenko, W.~Wang, T.~Weyand,
  M.~Andreetto, and H.~Adam.
\newblock Mobilenets: Efficient convolutional neural networks for mobile vision
  applications.
\newblock {\em ArXiv}, abs/1704.04861, 2017.

\bibitem{jain2007pores}
A.~K. {Jain}, Y.~{Chen}, and M.~{Demirkus}.
\newblock Pores and ridges: High-resolution fingerprint matching using level 3
  features.
\newblock {\em IEEE on Pattern Analysis and Machine Intelligence},
  29(1):15--27, 2007.

\bibitem{fvc2000}
D.~{Maio}, D.~{Maltoni}, R.~{Cappelli}, J.~L. {Wayman}, and A.~K. {Jain}.
\newblock Fvc2000: fingerprint verification competition.
\newblock {\em IEEE on Pattern Analysis and Machine Intelligence}, 2002.

\bibitem{fvc2002}
D.~{Maio}, D.~{Maltoni}, R.~{Cappelli}, J.~L. {Wayman}, and A.~K. {Jain}.
\newblock Fvc2002: Second fingerprint verification competition.
\newblock In {\em Object recognition supported by user interaction for service
  robots}, 2002.

\bibitem{fvc2004}
D.~Maio, D.~Maltoni, R.~Cappelli, J.~L. Wayman, and A.~K. Jain.
\newblock Fvc2004: Third fingerprint verification competition.
\newblock In D.~Zhang and A.~K. Jain, editors, {\em Biometric Authentication},
  pages 1--7. Springer Berlin Heidelberg, 2004.

\bibitem{marasco2014survey}
E.~Marasco and A.~Ross.
\newblock A survey on antispoofing schemes for fingerprint recognition systems.
\newblock {\em ACM Comput. Surv.}, 47(2), Nov. 2014.

\bibitem{marcialis2010analysis}
G.~L. {Marcialis}, F.~{Roli}, and A.~{Tidu}.
\newblock Analysis of fingerprint pores for vitality detection.
\newblock In {\em 2010 20th International Conference on Pattern Recognition},
  pages 1289--1292, 2010.

\bibitem{mura2015livdet}
V.~{Mura}, L.~{Ghiani}, G.~L. {Marcialis}, F.~{Roli}, D.~A. {Yambay}, and S.~A.
  {Schuckers}.
\newblock Livdet 2015 fingerprint liveness detection competition 2015.
\newblock In {\em IEEE 7th Intl. Conf. on Biometrics Theory, Applications and
  Systems (BTAS)}, pages 1--6, 2015.

\bibitem{livdet17}
V.~Mura, G.~Orrù, R.~Casula, A.~Sibiriu, G.~Loi, P.~Tuveri, L.~Ghiani, and
  G.~Marcialis.
\newblock Livdet 2017 fingerprint liveness detection competition 2017.
\newblock 03 2018.

\bibitem{fvc2006}
A.~F. R.~Cappelli, M.~Ferrara and D.~Maltoni.
\newblock Fingerprint verification competition 2006.
\newblock {\em Biometric Technology Today}, 15(7):7 -- 9, 2007.

\bibitem{Sandler2018MobileNetV2IR}
M.~{Sandler}, A.~{Howard}, M.~{Zhu}, A.~{Zhmoginov}, and L.~{Chen}.
\newblock Mobilenetv2: Inverted residuals and linear bottlenecks.
\newblock In {\em 2018 IEEE/CVF Conference on Computer Vision and Pattern
  Recognition}, pages 4510--4520, 2018.

\bibitem{Szegedy2016RethinkingTI}
C.~{Szegedy}, V.~{Vanhoucke}, S.~{Ioffe}, J.~{Shlens}, and Z.~{Wojna}.
\newblock Rethinking the inception architecture for computer vision.
\newblock In {\em 2016 IEEE Conference on Computer Vision and Pattern
  Recognition (CVPR)}, pages 2818--2826, 2016.

\bibitem{inception}
C.~Szegedy, V.~Vanhoucke, S.~Ioffe, J.~Shlens, and Z.~Wojna.
\newblock Rethinking the inception architecture for computer vision.
\newblock 06 2016.

\bibitem{ying2018liveface}
X.~Ying, X.~Li, and M.~C. Chuah.
\newblock Liveface: A multi-task cnn for fast face-authentication.
\newblock pages 955--960, 12 2018.

\end{thebibliography}
}

\clearpage
\appendix
\noindent \textbf{\Large{Appendix}}
\section{Back-propagation through the DHM} \label{sec:backprop_diagram}
\begin{figure}[h]
  \includegraphics[width=\columnwidth]{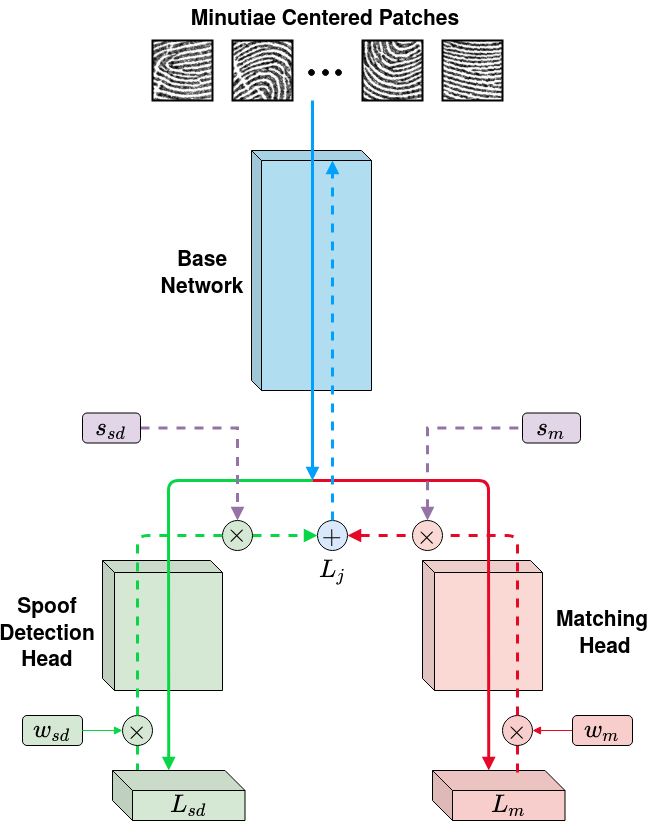}
  \caption{Back propagation through the DualHeadMobileNet}
  \label{fig:gradient_flow}
\end{figure}

As shown in Figure~\ref{fig:gradient_flow}, the spoof detection and matching heads learn only from their individual weighted gradients ($w_{sd}$ for spoof detection and $w_m$ for matching), while the base network learns from their sum to serve as a common feature extractor. $s_{sd}$ and $s_m$ are set to $-1$ if spoof detection or matching (respectively) are suppressed, otherwise both are set to $1$.

\section{Motivation: Architecture} \label{sec:motivation_architecture}
As explained in Section~\ref{sec:motivation} of the main text, we use a custom spoof detection network on top of features extracted by a fingerprint matching network. The spoof detection network (Table \ref{table:motivation_spoof_head}) takes feature maps of size 35 $\times$ 35 extracted from the $Mixed\_5d$ layer of an Inception-v3 fingerprint matching model (consisting of 1M parameters approx.) and consists of 700K parameters.
\vspace{-0.5mm}

\begin{table}[h]
\centering
\footnotesize
\caption{Custom Spoof Detection Head}
{
\begin{tabular}{  P{2.5cm} P{2cm} P{1.8cm}}
        \toprule
        \textbf{Layer} & \textbf{Kernel / Stride} & \textbf{Output size} \\[0.2cm]
        \midrule \midrule
        conv (groups = 288) & $3\times3 / 1$ & $36\times36\times288$ \\[0.2cm]
        \midrule
        conv & $1\times1 / 1$ & $36\times36\times512$ \\[0.2cm]
        \midrule
        conv (groups = 512) & $3\times3 / 2$ & $17\times17\times512$ \\[0.2cm]
        \midrule
        conv & $1\times1 / 1$ & $17\times17\times1024$ \\[0.2cm]
        \midrule
        conv (groups = 1024) & $3\times3 / 2$ & $8\times8\times1024$ \\[0.2cm]
        \midrule
        pool & $8\times8 / 8$ & $1\times1\times1024$ \\[0.2cm]
        \midrule
        flatten & & 1024 \\[0.2cm]
        \midrule
        linear & & 2\\[0.2cm]
        \bottomrule
\end{tabular}}
\label{table:motivation_spoof_head}
\vspace{-2mm}
\end{table}

\section{Spoof Detection Performance}
\label{sec:spoof_detection_long}

\begin{table*}[t]
\centering
\footnotesize
\caption{Spoof Detection Classification Errors \textsuperscript{$\dagger$}}
{\renewcommand{\arraystretch}{1}%
\begin{tabular}{ P{2cm} | P{3cm} | P{1.25cm} P{1.25cm} | P{1.25cm} P{1.25cm} | P{1.25cm} P{1.25cm} }
        \toprule
        
        \multirow{2}[5]{2cm}{\centering\textbf{Dataset}} & \multirow{2}[5]{3cm}{\centering\textbf{Sensor}} & \multicolumn{2}{P{2.5cm}|}{\textbf{BPCER}} & \multicolumn{2}{P{2.5cm}|}{\textbf{APCER}} &
        \multicolumn{2}{P{2.5cm}}{\textbf{ACER}}
        \\[0.2cm]
        
        \cmidrule(r){3-8}
        
        & & \textbf{$m$-FSB}  & \textbf{DHM} & \textbf{$m$-FSB} & \textbf{DHM} & \textbf{$m$-FSB} & \textbf{DHM} \\[0.2cm] 
        
        \midrule\midrule
        
        \multirow{6}{2cm}{\centering\textbf{LivDet 2015}} & Cross Match & 0.40 & \textbf{0.20} & 0.35 & 0.35 & 0.38 & \textbf{0.28} \\[0.2cm]
         & Digital Persona & 2.90 & \textbf{2.50} & \textbf{3.13} & 4.07 & \textbf{3.01} & 3.28 \\[0.2cm]
         & Green Bit & 0.40 & 0.40 & 0.60 & \textbf{0.47} & 0.50 & \textbf{0.44} \\[0.2cm]
         & Hi Scan & 0.80 & 0.80 & 3.27 & \textbf{2.74} & 2.04 & \textbf{1.77} \\[0.2cm]
        
        \midrule
        
        \multirow{3}{2cm}{\centering\textbf{LivDet 2017}} & Digital Persona & 3.25 & \textbf{3.07} & 3.77 & \textbf{3.72} & 3.51 & \textbf{3.40} \\[0.2cm]
         & Green Bit & \textbf{3.06} & 3.83 & 2.20 & \textbf{2.00} & \textbf{2.63} & 2.92 \\[0.2cm]
        
        \bottomrule
\end{tabular}}
\flushleft{\footnotesize{\quad \quad \quad \textsuperscript{$\dagger$} $m$-FSB refers to the Fingerprint Spoof Buster ~\cite{chugh2018fingerprint} baseline and DHM refers to the proposed method. \quad } }
\label{table: main_spoof_detection_long}
\end{table*}

In compliance with the ISO standards, to compare the spoof detection performance we report the following metrics in Table~\ref{table: main_spoof_detection_long}:

\begin{itemize}
    \item \textbf{Attack Presentation Classification Error Rate (APCER)}: Percentage of misclassified presentation attacks for a fixed threshold.
    \item \textbf{Bona Fide Presentation Classification Error Rate (BPCER)}: Percentage of misclassified bona fide presentations for a fixed threshold.
    \item \textbf{Average Classification Error Rate (ACER)}: (APCER + BPCER) / 2

\end{itemize}

The threshold in both cases is fixed to 0.5.

\section{Evaluation of Matching Feature Vectors}
\label{sec:matching_feature_eval}

\begin{figure}[h]
\centering
\begin{subfigure}{.46\columnwidth}
  \centering
  \includegraphics[width=\columnwidth]{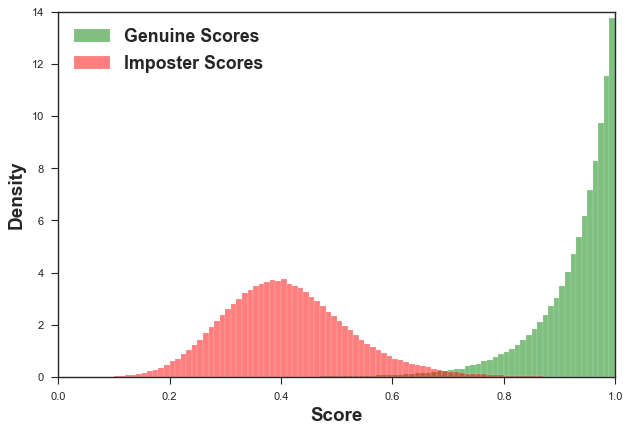}
  \caption{Baseline \cite{jain2018endtoend}}
  \label{fig:histogram:baseline}
\end{subfigure}
\hspace{1em}
\begin{subfigure}{.46\columnwidth}
  \centering
  \includegraphics[width=\columnwidth]{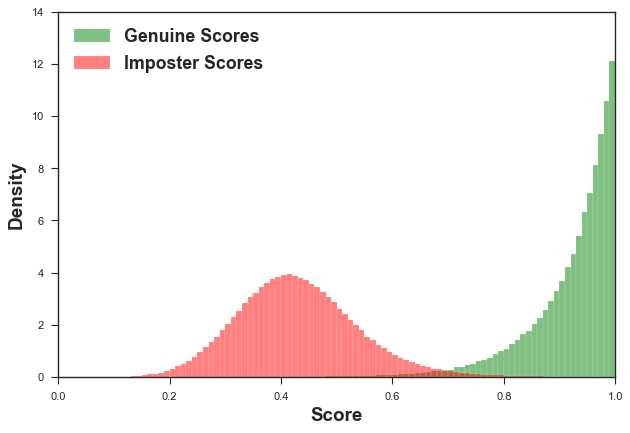}
  \caption{Ours}
  \label{fig:histogram:ours}
\end{subfigure}
\hspace{1em}
\caption{Histogram plot of the obtained matching scores. We can see a similar distribution of scores for both cases with a clear separation between genuine and imposter scores.}
\label{fig:histogram}
\end{figure}

Earlier, we compared the feature vectors extracted using our method with the baseline in terms of the fingerprint matching performance. Matching scores were obtained by average patch-level scores obtained from the extracted features, after removing false matches using a minutiae matcher \cite{jain2018endtoend}.
In this section, we directly compare the relevance of these extracted features by plotting the similarity scores for matching and non-matching pairs of patches. 

Since both of these methods are based on minutiae-centered patches and ground-truth minutiae correspondences are not available for the FVC datasets, we use VeriFinger to generate minutiae correspondences between matching fingerprint images. Minutiae-centered patches extracted around these corresponding minutiae points are treated as genuine or matching pairs. For imposter pairs we extract patches around any two non-matching minutiae points (belonging to non-matching fingerprint images). In total, we generate 380$K$ genuine or matching pairs and 380$K$ imposter or non-matching patch pairs. We pass these extracted patches through the network to obtain feature vectors and then calculate similarity scores using cosine distance. Figure~\ref{fig:histogram} shows a histogram of the obtained similarity scores. We can see a similar distribution of scores for the baseline \cite{jain2018endtoend} (left) and the proposed method (right) with a clear separation between genuine and imposter scores. We observe average similarity scores of 0.96 vs 0.96 (baseline vs ours) for matching pairs and 0.41 vs 0.43 (baseline vs ours) for imposter pairs.

\section{Robustness to Network Architecture} \label{sec:robustness_architecture}

In DHM we have used MobileNet-v2 \cite{Sandler2018MobileNetV2IR} as a starting point for our architecture. In order to examine the robustness of the proposed methodology to the underlying architecture, we also experiment with the popular ResNet-18 \cite{resnet} and Inception-v3 \cite{inception} networks and develop DualHeadResNet (DHR) and DualHeadInception (DHI) architectures analogous to DualHeadMobileNet. The architectures of DualHeadResNet (DHR) and DualHeadInception (DHI) are as follows:

\subsection{DHR}
We create the DHR network by splitting the ResNet-18 network \cite{resnet} at the $conv\_5x$ block and reducing the number of channels from 512 to 256 for that block (since our aim is to obtain a compact model for both the tasks) as shown in Table \ref{table:resnet}. This new model consists of approximately 6.4M parameters (4M out of which are part of the base network and are hence common for both spoof detection and matching) in comparison to 10.4M of a baseline system consisting of two separate ResNet-18 networks.

\begin{table}[h]
\footnotesize
\centering
\caption{Architecture of DHR}
\begin{tabular}{P{1.25cm}|P{2.25cm}|P{1.2cm}|P{2cm}}
    \toprule
    \textbf{Branch \newline{(Params)}} & \textbf{Type} & \textbf{Kernel / Stride} & \textbf{Input Size} \\
    \midrule
    \midrule
    \multirow{8}{1.25cm}{\centering{Base Network (4M)}}
                                  & conv padded & $7 \times 7$ / 2 & $224 \times 224 \times 3$  \\
                                  & maxpool padded & $3 \times 3$ / 2 & $112 \times 112 \times 64$  \\
                                  & conv\_2x* & - & $56 \times 56 \times 64$  \\
                                  & conv\_3x* & - & $56 \times 56 \times 64$  \\
                                  & conv\_4x* & - & $28 \times 28 \times 128$ \\ 
                                  & conv padded & $3 \times 3 / 2$ & $14 \times 14 \times 256$ \\
                                  & conv padded & $3 \times 3 / 1$ & $ 7 \times 7 \times 256$\\
                                  & conv & $1 \times 1 / 2$ & $ 7 \times 7 \times 256$\\
    \midrule
    \midrule
    \multirow{4}{1.4cm}{\centering{Spoof Detection Head (1.18M)}}
                            & conv padded & $3 \times 3 / 1$ & $ 4 \times 4 \times 256$\\
                            & conv padded & $3 \times 3 / 1$ & $ 4 \times 4 \times 256$\\
                            & avg pool & $4 \times 4$ & $ 4 \times 4 \times 256$\\
                            & linear $[256 \times 2]$  & - & $ 256$\\
                                
    \midrule
    \midrule
    \multirow{4}{1.4cm}{\centering{Matching Head (1.19M)}}
                        & conv padded & $3 \times 3 / 1$ & $ 4 \times 4 \times 256$\\
                            & conv padded & $3 \times 3 / 1$ & $ 4 \times 4 \times 256$\\
                            & avg pool & $4 \times 4$ & $ 4 \times 4 \times 256$\\
                            & linear $[256 \times 64]$  & - & $ 256$\\
    \bottomrule
\end{tabular}
\flushleft{\footnotesize{*ResNet blocks as defined in Table 1 of \cite{resnet}}.}
\vspace{-2mm}
\label{table:resnet}
\end{table}

\subsection{DHI}
Similarly, we create the DHI network by splitting the Inception-v3 network \cite{inception} after the \textit{Mixed\_7b} block as shown in Table \ref{table:inception}. This new model consists of approximately 28M parameters (15.2M out of which are part of the base network and are hence common for both spoof detection and matching) in comparison to 43.7M of a baseline system consisting of two separate Inception-v3 networks, while maintaining the accuracy for both the tasks. 

\begin{table}[h]
\footnotesize
\centering
\caption{Architecture of DHI}
\begin{tabular}{P{1.5cm}|P{2.25cm}|P{1.2cm}|P{2cm}}
    \toprule
    \textbf{Branch \newline{(Params)}} & \textbf{Type} & \textbf{Kernel / Stride} & \textbf{Input Size} \\
    \midrule
    \midrule
    \multirow{10}{1.4cm}{\centering{Base Network (15.2M)}}
                                  & conv & $3 \times 3$ / 3 & $448 \times 448 \times 3$  \\
                                  & conv & $3 \times3 $ / 1 & $149 \times 149 \times 32$ \\
                                  & conv padded & $3 \times 3$ / 1 & $147 \times 147 \times 32$ \\
                                  & pool & $3 \times 3$ / 2 & $147 \times 147 \times 64$ \\
                                  & conv & $1 \times 1$ / 1 & $73 \times 73 \times 64$ \\
                                  & conv & $3 \times 3$ / 1 & $73 \times 73 \times 80$ \\
                                  & pool & $3 \times 3$ / 2 & $71 \times 71 \times 192$ \\
                                  & 3 $\times$ Inception A* & - & $35 \times 35 \times 192$ \\
                                  & 5 $\times$ Inception B* & - & $35 \times 35 \times 288$ \\
                                  & 2 $\times$ Inception C* & - & $17 \times 17 \times 768$ \\
    \midrule
    \midrule
    \multirow{4}{1.4cm}{\centering{Spoof Detection Head (6M)}}
                                & Inception C* & - & $8 \times 8 \times 2048$ \\
                                & pool & $8 \times 8$ & $8 \times 8 \times 2048$ \\
                                & linear $[2048 \times 2]$ & - & $1 \times 1 \times 2048$ \\
                                & softmax & - & $2$ \\
                                
    \midrule
    \midrule
    \multirow{3}{1.4cm}{\centering{Matching Head (6.2M)}}
                                & Inception C* & - & $8 \times 8 \times 2048$ \\
                                & pool & $8 \times 8$ & $8 \times 8 \times 2048$ \\
                                & linear $[2048 \times 64]$ & - & $2048$ \\
    \bottomrule
\end{tabular}
\flushleft{\footnotesize{*Inception A, Inception B and Inception C refer to the three types of Inception blocks defined in \cite{Szegedy2016RethinkingTI}.}}
\label{table:inception}
\vspace{-2mm}
\end{table}

\begin{table}[t]
\centering
\footnotesize
\caption{Performance comparison of the DHR and DHI networks: proposed method is robust to the backbone architecture}
{\renewcommand{\arraystretch}{1}%
\begin{tabular}{  P{2cm} P{1.5cm} P{1.5cm} P{1.5cm}  }
        \toprule
        
        \multicolumn{4}{P{7cm}}{\textbf{Matching Performance (FRR (\%) @ FAR = 0.1\%)}} \\[0.2cm]
        
        \midrule
        
        \textbf{Method} & \textbf{FVC 2006 DB2A} & \textbf{FVC 2004 DB1A} & \textbf{FVC 2004 DB2A} \\[0.2cm]
        
        \midrule
        
        DHM & 0.00 & 5.89 & 6.99 \\[0.1cm]
        
        \midrule
        
        DHR & 0.00 & 5.54 & 5.81 \\[0.1cm]
        
        \midrule
        
        DHI & 0.02 & 5.43 & 6.46 \\[0.1cm]

        \midrule\midrule
        
        \multicolumn{4}{P{7cm}}{\textbf{Spoof Detection Perf. (LivDet-15 CrossMatch sensor)}} \\
        
        \midrule
        
        \textbf{Method} & \textbf{ACE (\%)} & \multicolumn{2}{P{3.5cm}}{ \textbf{E\textsubscript{$fake$} @ E\textsubscript{$live$} = 0.1\%}} \\[0.2cm]
        
        \midrule
        
        ResNet-18 \cite{resnet} & 0.5	& \multicolumn{2}{P{3cm}}{0.9} \\[0.1cm]
        
        \midrule
        
        DHR & 0.48 & \multicolumn{2}{P{3cm}}{0.69} \\[0.1cm]
        
        \midrule
        
        Inception-v3 \cite{Szegedy2016RethinkingTI} & 0.44 & \multicolumn{2}{P{3cm}}{0.55} \\[0.1cm]
        
        \midrule
        
        DHI & 0.28 & \multicolumn{2}{P{3cm}}{0.76} \\[0.1cm]
        
        \bottomrule
\end{tabular}}
\label{table: arch_res}
\end{table}

\quad

As shown in Table \ref{table: arch_res}, changing the underlying architecture has little effect on both the spoof detection and matching performance. 

\section{Implementation Details} \label{sec:hyperparams}
Input images are first processed (as done by \cite{Gajawada2019UniversalMT}) to crop out the region of interest from the background, followed by extraction of minutiae-centered and oriented patches (using VeriFinger SDK) of size 96 $\times$ 96 and their descriptors using \cite{jain2018endtoend}. We only used patches and descriptors of a single-type (minutiae-centered, size 96 $\times$ 96) instead of an ensemble of descriptors since (i) we did not observe much boost in the matching performance on using multiple descriptors and (ii) using an ensemble slows down the system considerably which goes against our motivation. For training, 20\% of the data for each sensor was used for validation for the ReduceLROnPlateau scheduler (initial learning rate of $10^{-3}$ with patience 10 until $10^{-8}$) and Adam optimiser. Weights for the spoof detection and matching loss, i.e, $w_{sd}$ and $w_m$ were set to 1 and 10 respectively according to the order of the losses. For inference we have used the minutiae matching algorithm \cite{jain2018endtoend} to aggregate and compare the descriptors extracted by our network to obtain similarity scores between two fingerprints. All experiments have been performed using PyTorch framework and two Nvidia 2080-Ti GPUs.

\end{document}